\documentclass[11pt]{article}
 
\usepackage[preprint]{acl}
 
\usepackage{times}
\usepackage{latexsym}
\usepackage[T1]{fontenc}
\usepackage[utf8]{inputenc}
\usepackage{microtype}
\usepackage{inconsolata}
\usepackage{graphicx}
\usepackage{booktabs}
\usepackage{float}
\usepackage{multirow}

\usepackage[dvipsnames]{xcolor}
\usepackage{pgf}
\usepackage{colortbl}

\newcommand{\gradientcell}[6]{%
    \def\value{#1}%
    \def\minvalue{#2}%
    \def\maxvalue{#3}%
    \def\mincolor{#4}%
    \def\maxcolor{#5}%
    \def\transparency{#6}%
    \ifdimcomp{\value pt}{>}{\maxvalue pt}{\cellcolor{#5!100.0!#4!#6}\value}{%
    \ifdimcomp{\value pt}{<}{\minvalue pt}{\cellcolor{#5!0.0!#4!#6}\value}{%
         \pgfmathparse{int(round(100*(#1/(\maxvalue-\minvalue))-(\minvalue *(100/(\maxvalue-\minvalue)))))}%
        \xdef\tempa{\pgfmathresult}%
        \cellcolor{#5!\tempa!#4!#6}\value%
    }}%
}

\definecolor{LightGray}{gray}{0.95}
\newcommand{\R}[1]{\gradientcell{#1}{.0}{.9}{LightGray}{Goldenrod}{60}}

\newcommand{\D}[2]{%
    \pgfmathparse{max(0,min(100,int(round(100*(#2/(.9-.0))-(.0*(100/(.9-.0)))))))}%
    \xdef\mainpct{\pgfmathresult}%
    \cellcolor{Goldenrod!\mainpct!LightGray!60}%
    \pgfmathparse{max(0,min(100,int(round(100*(#1/(.9-.0))-(.0*(100/(.9-.0)))))))}%
    \xdef\prefixpct{\pgfmathresult}%
    {\tiny\setlength{\fboxsep}{0.2pt}\colorbox{Goldenrod!\prefixpct!LightGray}{#1}$\rangle$~}%
    #2%
}
 
\title{CHALIS: A Challenge Dataset for Language Identification\\in Difficult Scenarios}
 
\author{Michal Tichý \and Jindřich Libovický \\
 Charles University, Faculty of Mathematics and Physics \\
 Institute of Formal and Applied Linguistics \\
 V Holešovičkách 747/2, 180 00 Praha, Czechia \\
  \texttt{libovicky@ufal.mff.cuni.cz}}
 
\begin{document}
\maketitle
 
\begin{abstract}
We present \textbf{CHALIS} (Challenging Language Identification Samples), a new benchmark dataset explicitly designed to address difficult cases in language identification: cousin languages and orthographic noise.
Our dataset has two parts: First, we collected sentences shared across mutually intelligible language pairs (Czech/Slovak, Spanish/Catalan, Portuguese/Galician, Danish/Norwegian).
The second part tests for orthography noise: we transliterate text across multiple scripts, remove diacritics, simulate homoglyph attacks, and use Internet slang.
We evaluate four widely used language identification systems on CHALIS and demonstrate that all struggle substantially in these scenarios, especially on lower-resource languages within cousin pairs and on transliterated input.
%
The resource is publicly available at \url{https://huggingface.co/datasets/michal-tichy/CHALIS}.
\end{abstract}
 
\section{Introduction}
 
Language identification (LangID), the task of determining the language of a given text, is a foundational component of many NLP pipelines \citep{langIdSurvey}.
Although some researchers have characterized it as a solved problem \citep{langIdSolved}, growing demand for large-scale web-crawled multilingual corpora \citep{penedo2025fineweb,burchell-etal-2025-expanded} has exposed significant weaknesses in current systems.
 
\citet{qualityAtGlance} audited 205 languages across five datasets and found that over 8\% of samples were mislabeled with the wrong language, and a further 10\% were not linguistic in character at all. The most error-prone cases involve languages that are closely related to a higher-resource neighbor, and texts that have been transliterated into a different script.
 
These findings motivate our targeted evaluation benchmark. Existing datasets used to assess LangID models, such as FLORES-200 \citep{flores}, originally designed for multilingual machine translation, do not specifically probe these hard cases. 

We introduce CHALIS to fill this gap. The dataset covers both \textbf{cousin language pairs} and \textbf{transliteration and orthographic variation}, with a human-annotated subset for cousin language pairs where annotators validated whether sentences are grammatically correct in each language. We evaluate four state-of-the-art LangID systems on CHALIS, revealing systematic failure modes in both scenarios.
 
\section{Background}
 
\subsection{Language Identification}
 
Language identification is a classic text classification task \citep{langIdSurvey}. Modern systems such as FastText \citep{fastTextClassification}, OpenLID \citep{openLID}, GlotLID \citep{glotlid}, and GCLD3 achieve strong results on standard benchmarks. However, as \citet{langIdInTheWild} shows, performance degrades significantly in realistic web-crawled settings, where mislabeled data is common.
 
\subsection{Cousin Languages}
\label{sec:cousins}
 
Defining a ``language'' is a notoriously difficult exercise, nicely illustrated by the Weinreich witticism: \textit{``a language is a dialect with an army and navy''} \citep{weinreichWitt}. Language communities that coexist geographically develop varieties that mutually influence each other, often remaining \emph{mutually intelligible} \citep{dialectology}. Examples include Norwegian and Danish, or Czech and Slovak.
 
We refer to such pairs as \textbf{language cousins}. They pose a dual challenge for LangID: they are not equally represented in training data (one is typically less prevalent), and there exist sentences that are grammatically correct in \emph{both} languages. For example, \textit{``Je to obrovský problém''} (``It is a huge problem'') is valid in both Czech and Slovak. Both \citet{qualityAtGlance} and \citet{langIdInTheWild} note that mistaking a lower-resource language for its more prominent cousin is among the most frequent errors in web-crawled corpora.
 
\subsection{Non-standard Orthography}
 
Transliteration, i.e., converting text from one script to another, often approximating pronunciation, is another major source of LangID errors. Transliterated text does not match the script profile that LangID models expect, yet it retains the language's lexical and syntactic patterns. For instance, Russian Cyrillic transliterated to Latin characters is meaningless to a model that associates Russian exclusively with the Cyrillic script. Moreover, different languages have different transliteration standards.

The most frequent orthography issues in writing are probably mistakes and typos. These naturally occur in most digital texts and presumably also in datasets used for LangID training. However, there are more systematic issues that might affect the LangID performance.
These issues include removing diacritics and only using ASCII characters in Latin, historically driven by software and hardware limitations. Homoglyph attacks use the fact that some graphemes look similar across scripts (e.g., o and o in Latin and Cyrillic) and are often used to confuse automatic systems such as spam detectors. 
 
\section{The CHALIS Dataset}

We collect a challenge set that addresses the issues discussed in the previous section. The dataset consists of three parts: Sentences that are valid in two cousin languages (and multiclass classifiers need to decide for one of them only), sentences that are only valid in one of the cousin-language pairs, but consist of words that are common in both languages, and the third part focuses on the orthography phenomena, and unlike the first two, it was created fully automatically.



Basic statistics of the dataset are presented in Table~\ref{tab:stats}. The dataset is available at Huggingface Hub\footnote{\url{https://huggingface.co/datasets/michal-tichy/CHALIS}} under the CC-BY-NC 4.0 license, which is consistent with the underlying data. More technical details are in Appendix~\ref{app:format}.

\begin{table}[t]
  \centering\small
  \begin{tabular}{llcc}
    \toprule
    & \textbf{Subset / Language} & \textbf{Lines} & \textbf{Avg.\ Words} \\
    \midrule
    \multirow{4}{*}{\rotatebox{90}{Joint}}
                   & Czech \& Slovak      & 655   & 5.9  \\
                   & Spanish \& Catalan   & 642   & 5.8  \\
                   & Danish \& Norwegian  & 459   & 7.0  \\
                   & Portuguese \& Galician & 1,143 & 7.9 \\
    \midrule
    \multirow{4}{*}{\rotatebox{90}{Single}}
                   & Czech                & 491   & 6.5  \\
                   & Danish               & 361   & 8.7  \\
                   & Norwegian            & 327   & 7.9  \\ 
                   & Spanish              & 546   & 6.5  \\
    \midrule
    \multirow{7}{*}{\rotatebox{90}{Ortography}}
                   & Cyrilics to English     & 50k & 15.4 \\
                   & Cyrilics to Czech       & 10k & 13.9 \\
                   & Arabic (Kazakh)      & 10k & 12.0 \\
                   & Diacritics           & 30k & 17.2 \\
                   & Homoglyps            & 30k & 14.5 \\
                   & Antspeak             & 10k & 11.8 \\
                   & Leetspeak            & 10k & 14.3 \\
    \bottomrule
  \end{tabular}
  
  \caption{CHALIS dataset statistics.}
  \label{tab:stats}
\end{table}
 
\subsection{Cousin Language Data}
\label{sec:cousin-data}
 
We collected data for four cousin pairs, all Indo-European languages from Europe. Czech/Slovak (Slavic), Spanish/Catalan (Romance), Portuguese/Galician (Romance), Danish/Norwegian (Germanic). There is a different level of resource disparities between the languages. Czech and Slovak are both official languages of the countries where they are spoken, with roughly 10M and 5M speakers respectively. There is a similar situation with Danish and Norwegian, both of which are spoken by roughly 5M people. The situation is slightly different with the Romance languages. Due to the colonial history, Spanish is spoken by 500M speakers, whereas Catalan is spoken by only 10M people in Europe. Similarly, Portuguese is spoken by 250M people in Europe and South America, whereas Galician is only spoken by around 3M people in Spain.
 
\paragraph{Candidate sentence retrieval.}

For each language in a pair, we tokenized text using Moses tokenizer \citep{koehn-etal-2007-moses} and built a frequency dictionary from the WMT Monolingual News Crawl \citep{wmtOverview}, supplemented by Wikipedia-sourced data from the Leipzig Corpora Collection \citep{leipzig_collection} for lower-resource languages. We extracted the 10,000 most frequent words in each language and computed their intersection to obtain a shared vocabulary. Table~\ref{tab:vocab} shows the resulting shared vocabulary sizes.
 
\begin{table}[t]
  \centering\footnotesize
  \begin{tabular}{lcc}
    \toprule
     &  & \textbf{Count} \\
    \midrule
    Czech       & Slovak    & 2,339 \\
    Danish      & Norwegian & 4,266 \\
    Spanish     & Catalan   & 1,623 \\
    Portuguese  & Galician  & 3,491 \\
    \bottomrule
  \end{tabular}
  
  \caption{Shared vocabulary sizes per language pair (out of 10k must frequent tokens in the languages).}
  \label{tab:vocab}
\end{table}
 
We then retrieved all sentences from the source corpora whose tokens all appear in the shared vocabulary, yielding candidate sentences that \emph{could} belong to either language.
 
\paragraph{KenLM scoring.}

Candidates were scored using 4-gram KenLM language models \citep{kenlmPaper} with Kneser–Ney smoothing \citep{kneser-ney}, trained on the source corpora. Each sentence received a probability under each language's model, normalized for length. We ranked sentences by the harmonic mean of the two normalized log-likelihoods, and selected the top 1,200 per pair as most plausibly correct in both languages.
 
\paragraph{Named entity filtering.}

Inspection of high-scoring sentences revealed that many consisted largely of proper names (e.g., sports results). We applied the GLiNER named-entity recognizer \citep{gliner} to remove sentences in which more than 25\% of tokens were classified as named entities (persons, locations, organizations, etc.). Near-duplicate sentences were also removed using an edit-distance threshold of 10\% of sentence length.
 
\paragraph{Human annotation.}

The top 1,200 filtered sentences per pair were presented to human annotators, who were native speakers or advanced students of the target languages.
Annotators labeled each sentence as valid or invalid in their language. Where two annotators were available (Czech, Slovak, Danish), only sentences with unanimous agreement were kept. More details on the annotation process are in Appendix~\ref{app:annotation}.

The resulting items fall into two subsets:
\begin{itemize}
  \item Single: sentences valid in \emph{only one} language but structurally similar to the other—a particularly hard case for LangID. We disregard language, for which only a few sentences remained after filtering.
  \item Joint: sentences valid in \emph{both} languages of the pair.
\end{itemize}

We evaluate the `Single` portion of the dataset with the F1 score. For the `Joint` part of the dataset, we report the distribution of how often the model chose the respective language in language pairs vs. other languages.
 
\subsection{Transliteration Data}
 
\paragraph{Cyrillic to Latin.}
We transliterated Bulgarian, Macedonian, Russian, Serbian, and Ukrainian text (sourced from the Leipzig Wikipedia corpus) into Latin script using two target typographies: English (via CyrTranslit; \citealp{cyrtranslit}). We further transliterated Russian into Czech using Czech orthography (via conversion scripts from the Charles translator project; \citealp{charles_translator}).
 
\paragraph{Kazakh to Arabic.}

Kazakh has historically been written in Arabic script (before Soviet-era Cyrillisation), and standards for Arabic-script Kazakh are well established. We transliterated Cyrillic Kazakh text into Arabic using scripts from the MC\textsuperscript{2} dataset \citep{mc2}.

\paragraph{Homoglyph attacs.}

In the Latin text, letters are randomly replaced by visually identical Cyrillic or Greek characters in Czech, Slovak, and English.

\paragraph{Internet slang.}

We also include several related phenomena: (1) \textbf{diacritics removal} for Czech, Slovak, and Vietnamese; (2) \textbf{antspeak} (\texttt{s} \texttt{p} \texttt{a} \texttt{c} \texttt{e} \texttt{s} \texttt{b} \texttt{e} \texttt{t} \texttt{w} \texttt{e} \texttt{e} \texttt{n} \texttt{l} \texttt{e} \texttt{t} \texttt{t} \texttt{e} \texttt{r} \texttt{s}) and \textbf{leetspeak} (character substitution, e.g.\ \texttt{n00b}) for English, using pyleetspeak \citep{pyleetspeak}.

\section{Evaluation}
 
\subsection{Models}
 
We evaluate four widely used LangID systems: \textbf{FastText} \citep{fastTextClassification}, \textbf{OpenLID} \citep{openLID}, \textbf{GlotLID} \citep{glotlid}, and \textbf{GCLD3}. All four are multiclass classifiers that always output exactly one language label. More details on the models are in Table~\ref{tab:modelsinfo} in Appendix~\ref{app:software}.

\begin{table*}[ht]
\centering
\small
\setlength{\tabcolsep}{5pt}
\begin{tabular}{l
  c|
  rr
  rr
  @{\hspace{6pt}}
  |@{\hspace{6pt}}
  rrr|
  rrr|
  rrr|
  rrr
}
\toprule
& \multirow{2}{*}{\shortstack{FLORES\\200}}
& \multicolumn{4}{c}{Single (F1 score)}
& \multicolumn{12}{c}{Joint (distribution)} \\
\cmidrule(lr{8pt}){3-6}\cmidrule(l{-4pt}r){7-18}
\cmidrule(lr){7-9}\cmidrule(lr){10-12}\cmidrule(lr){13-15}\cmidrule(lr){16-18}
& & ces & spa
& nob & dan
& ces & slk & oth
& spa & cat & oth
& por & glg & oth
& nob & dan & oth \\
\midrule
FastText
  & \R{.912}
  & \R{.84} & \R{.93} & \R{.87} & \R{.90}
  & \R{.38} & \R{.62} & \R{.00} & \R{.76} & \R{.15} & \R{.09} & \R{.85} & \R{.15} & \R{.00} & \R{.56} & \R{.43} & \R{.01} \\
OpenLID
  & \R{.926}
  & \R{.95} & \R{.96} & \R{.79} & \R{.82}
  & \R{.91} & \R{.09} & \R{.00} & \R{.72} & \R{.01} & \R{.27} & \R{.68} & \R{.31} & \R{.01} & \R{.67} & \R{.29} & \R{.04} \\
GlotLID
  & \R{.939}
  & \R{.95} & \R{.94} & \R{.88} & \R{.90}
  & \R{.83} & \R{.17} & \R{.00} & \R{.81} & \R{.12} & \R{.07} & \R{.97} & \R{.02} & \R{.01} & \R{.57} & \R{.43} & \R{.00} \\
GCLD3
  & \R{.352}
  & \R{.71} & \R{.84} & \R{.76} & \R{.83}
  & \R{.71} & \R{.21} & \R{.08} & \R{.55} & \R{.34} & \R{.11} & \R{.50} & \R{.47} & \R{.03} & \R{.44} & \R{.50} & \R{.06} \\
\bottomrule
\end{tabular}
\caption{Language identification results for cousin language pairs under single and joint prediction.
  \emph{oth} = fraction assigned to neither language in the pair. The low score of GCLD3 on the FLORES dataset is due to the smaller language coverage of the system.}
\label{tab:langid}
\end{table*}

\begin{table*}
\centering\footnotesize
\setlength{\tabcolsep}{2pt}

\begin{tabular}{l
  rrrr|
  r|
  r
  @{\hspace{6pt}}
  |@{\hspace{6pt}}
  rrr|
  rrr|
  rr
}
\toprule
& \multicolumn{6}{c}{Transliteration to}
& \multicolumn{8}{c}{Orthography noise} \\
\cmidrule(lr{8pt}){2-7}\cmidrule(l{-4pt}r){8-15}
& \multicolumn{4}{c}{eng}
& ces
& arb
& \multicolumn{3}{c}{Diacritics}
& \multicolumn{3}{c}{Homoglyphs}
& \multicolumn{2}{c}{Internet slang} \\
\cmidrule(lr){2-5}\cmidrule(lr){6-6}\cmidrule(lr){7-7}
\cmidrule(lr){8-10}\cmidrule(lr){11-13}\cmidrule(lr){14-15}
& \multicolumn{1}{c}{bul} & \multicolumn{1}{c}{mkd} & \multicolumn{1}{c}{rus} & \multicolumn{1}{c}{srp}
& \centering rus
& \centering kaz
& \multicolumn{1}{c}{ces} & \multicolumn{1}{c}{slk} & \multicolumn{1}{c}{vie}
& \multicolumn{1}{c}{eng} & \multicolumn{1}{c}{ces} & \multicolumn{1}{c}{slk}
& \multicolumn{1}{c}{Ant} & \multicolumn{1}{c}{Leet} \\
\midrule

FastText
  & \D{.99}{.00} & \D{.99}{.00} & \D{.99}{.00} & \D{.99}{.00}
  & \D{.99}{.00}
  & \D{.99}{.00}
  & \D{1}{.89} & \D{1}{.92} & \D{1}{.90}
  & \D{.99}{.92} & \D{1}{.98} & \D{1}{.97}
  & \D{.99}{.12} & \D{.99}{.98} \\
OpenLID
  & \D{.99}{.00} & \D{.99}{.00} & \D{.95}{.00} & \D{.99}{.01}
  & \D{.95}{.00}
  & \D{.99}{.00}
  & \D{.99}{.81} & \D{.99}{.88} & \D{.99}{.94}
  & \D{.99}{.54} & \D{.95}{.96} & \D{.99}{.94}
  & \D{.95}{.00} & \D{.95}{.91} \\
GlotLID
  & \D{1}{.00} & \D{1}{.00} & \D{1}{.00} & \D{1}{.00}
  & \D{1}{.00}
  & \D{1}{.00}
  & \D{1}{.85} & \D{1}{.91} & \D{1}{.89}
  & \D{.99}{.77} & \D{1}{.97} & \D{1}{.97}
  & \D{.99}{.00} & \D{.99}{.92} \\
GCLD3
  & \D{.97}{.68} & \D{.97}{.00} & \D{.98}{.76} & \D{.97}{.00}
  & \D{.98}{.48}
  & \D{.99}{.00}
  & \D{.97}{.62} & \D{.97}{.74} & \D{.98}{.13}
  & \D{.93}{.35} & \D{.97}{.80} & \D{.97}{.72}
  & \D{.93}{.00} & \D{.93}{.72} \\

\bottomrule
\end{tabular}

\caption{Language identification results (F1 score) for transliteration and orthography noise part of the dataset. Each cell contains the F1 score before applying the orthography noise in a smaller font, followed by $\rangle$ sign and the F1 score on the orthography modification of the dataset.}
 
\end{table*}`

\subsection{Results on Cousin Languages}

Table~\ref{tab:langid} shows the results both for the Single-language subsect (evaluation using F1 score) and the language distributions for the Joint subset. In the `Single' subset, the scores are relatively low, given that they are high-resource languages. They lag behind the models' overall performance, indicating they are easy to confuse with the cousin languages.

The `Joint' subset (sentences valid in both languages) reveals strong systematic bias. For Czech/Slovak, OpenLID and GlotLID assign over 90\% of jointly valid sentences to Czech, effectively ignoring Slovak. Surprisingly, FastText prefers lower-resourced Slovak. A similar pattern holds for Portuguese/Galician and Spanish/Catalan. GCLD3 shows less extreme bias but achieves lower overall F1, suggesting it trades accuracy for a more balanced, if still imprecise, distribution.

\subsection{Results on Orthography Issues}

Table~2 presents the results on the orography subset of the dataset. The F1 scores of all systems before the orthography modification are almost perfect, showing that the work with examples that are otherwise easy for LangID systems.

Transliteration is catastrophic for nearly all systems. FastText, OpenLID, and GlotLID score 0 across all Cyrillic-to-Latin conditions. They fail to recognize the language at all once the expected script is removed. GCLD3 shows partial robustness for Russian and Bulgarian transliteration into English (F1 scores of 0.76 and 0.68), though it, too, fails on Macedonian and Serbian, and, while not entirely, on Czech-orthography Russian. All systems score 0 on Arabic-script Kazakh.

In contrast, diacritics removal and homoglyph substitution have a more moderate, but still strong impact. Most systems retain F1 above 0.80 for diacritics-stripped Czech, Slovak, and Vietnamese, and above 0.9 for Czech and Slovak homoglyph attacks. OpenLID is a partial exception, dropping to 0.54 on English homoglyphs.

Antspeak proves most disruptive: FastText scores 0.12, while all other systems score 0, suggesting that inter-character spacing breaks n-gram statistics entirely. Leetspeak is handled more gracefully, with all systems achieving F1 above 0.70.

\section{Conclusions}

We presented CHALIS, a challenge benchmark targeting two systematic failure modes in language identification: cousin-language discrimination and non-standard orthography. Evaluating four widely used LangID systems, we found a consistent asymmetry: models reliably identify higher-resource languages while largely failing on lower-resource counterparts, with several models assigning nearly all ambiguous sentences to the dominant language in a pair. Transliteration proves catastrophic across the board, while antspeak similarly defeats current approaches, and diacritics removal and homoglyph substitution are handled more gracefully. These findings suggest that current LangID systems overfit to surface-level script and frequency cues. We hope CHALIS will motivate the development of stronger identification systems for lower-resource languages whose speakers deserve reliable NLP tooling.

\section*{Limitations}

CHALIS covers only four cousin language pairs, all Indo-European languages from Europe. While these pairs illustrate systematic LangID failure modes, the benchmark does not capture the full diversity of closely related language families, such as Arabic dialects, Turkic languages, or South and Southeast Asian language groups, where resource disparities and mutual intelligibility raise similar concerns.

The human annotation phase, which validates the `Single' and `Joint' subsets, was conducted by a small number of annotators. 
The orthography subset was constructed fully automatically and does not reflect the natural distribution of transliteration or slang usage found in real web-crawled data. Transliteration standards vary across communities and over time, and our choices (e.g., CyrTranslit for Cyrillic-to-English, Czech orthography for Russian) represent only a subset of conventions in actual use. Similarly, antspeak and leetspeak were generated programmatically rather than sampled from genuine user-generated content.

Finally, CHALIS is an evaluation-only benchmark. It does not include a training signal, so it cannot be used directly to improve LangID systems; it can only be used to diagnose them. Addressing the failure modes we identify will require targeted data collection and model development efforts beyond the scope of this work.

\section*{Miscellaneous}

We used GitHub Copilot when writing code. We used Grammarly for grammar checking, Claude for proofreading and for text shortening.
 
 
 
\bibliography{custom}

\appendix

\begin{table*}[t]
\centering\small
\begin{tabular}{llll}
\toprule
\textbf{Model} & \textbf{Architecture} & \textbf{Model size} & \textbf{License} \\
\midrule
FastText (lid218e) & FastText & 1.2 GB & CC BY-NC 4.0 \\
OpenLID            & FastText & 1.3 GB & GPL 3.0        \\
GlotLID (v3)       & FastText & 1.7 GB & Apache 2.0     \\
GCLD3              & Char-n-gram-based NN & 15MB & Apache 2.0 \\
\bottomrule
\end{tabular}
\caption{Overview of evaluated LangID systems. Model sizes are taken from the \texttt{.bin} file, where available. FastText-based models do not report discrete parameter counts.}
\label{tab:modelsinfo}
\end{table*}

\section{Annotation Procedure Details}\label{app:annotation}

Annotators were university students recruited via internal social media. We explained to them what data we want to annotate for what purpose. The annotators were asked to decide whether a sentence is in their native language. We estimated that processing 1,200 sentences would take approximately one hour and offered the students a compensation of 500 CZK (3.7$\times$ the minimum wage and 2$\times$ the median wage in 2025 in Czechia).

\section{Licenses of Software and Data}\label{app:software}

Table~\ref{tab:modelsinfo} summarizes the licenses of the four evaluated LangID systems. All evaluated models are freely available for research use: FastText (lid218e) is released under CC BY-NC 4.0, restricting commercial use; OpenLID under GPL 3.0; and both GlotLID and GCLD3 under the permissive Apache License 2.0. The training corpora used to construct CHALIS are similarly open: WMT News Crawl data and the Leipzig Corpora Collection are freely available for research purposes, and Wikipedia-derived data is released under CC BY-SA 4.0. The KenLM language modeling toolkit is distributed under the LGPL 3.0 license. The GLiNER named-entity recognizer used for filtering is released under the Apache License 2.0. Transliteration tools include CyrTranslit (MIT license), the Charles Translator conversion scripts, and the MC\textsuperscript{2} dataset scripts; pyleetspeak is also MIT-licensed. The Moses tokenizer is distributed under the LGPL 2.1 license. CHALIS itself will be released under CC BY 4.0 to maximize reusability.

\section{Dataset Format}\label{app:format}

CHALIS is distributed as a single JSON file structured as a dictionary with the following top-level keys: \texttt{single}, \texttt{joint}, \texttt{cyrilic\_to\_czech}, \texttt{cyrilic\_to\_english}, \texttt{antspeak}, \texttt{arabic}, \texttt{diacritics}, \texttt{homoglyphs}, and \texttt{leet}. Each key maps to a list of items, where every item is a dictionary with the following fields:

\begin{itemize}
    \item \texttt{text}: the string to be classified;
    \item \texttt{label}: a list of language labels in BCP-47-style format extended with script information (e.g., \texttt{['ces\_Latn', 'slk\_Latn']}), following the OpenLID convention. For the \texttt{joint} subset, the list contains two labels; for all other subsets, it contains exactly one;
    \item \texttt{original}: the source text prior to rule-based modification. This field is present in all subsets except \texttt{single} and \texttt{joint}, where the texts were collected directly rather than derived from another text.
\end{itemize}

\end{document}